\documentclass[10pt,twocolumn,letterpaper]{article}

\usepackage{wacv}
\usepackage{times}
\usepackage{epsfig}
\usepackage{graphicx}
\usepackage{amsmath}
\usepackage{amssymb}
\usepackage{bbold}
\usepackage{authblk}
\newcommand{\myparagraph}[1]{\vspace{3pt}\noindent{\bf #1}}
\usepackage[ruled,linesnumbered]{algorithm2e}
\newcommand{\squeezeup}{\vspace{-2.5mm}}

\wacvfinalcopy 

\def\wacvPaperID{1216} 

\ifwacvfinal\fi
\begin{document}

\title{
Understanding Misclassifications by Attributes 
\vspace{-5mm}}

\author[1]{Sadaf Gulshad}
\author[1]{Zeynep Akata}
\author[2]{Jan Hendrik Metzen}
\author[1]{Arnold Smeulders}
\affil[1]{UvA-Bosch Delta Lab\\
University of Amsterdam, The Netherlands\\}
\affil[2]{Bosch Center for AI (BCAI), Renningen, Germany}

\maketitle
\begin{abstract}

In this paper, we aim to understand and explain the decisions of deep neural networks by studying the behavior of predicted attributes when adversarial examples are introduced.  We study the changes in attributes for clean as  well  as  adversarial  images in both standard and adversarially robust networks. We propose a metric to quantify the robustness of an adversarially robust network against adversarial attacks. In a standard network, attributes predicted for adversarial images are consistent with the wrong class, while attributes predicted for the clean images are consistent with the true class. In an adversarially robust network, the attributes predicted for adversarial images classified correctly are consistent with the true class. Finally, we show that the ability to robustify a network varies for different datasets. For the fine grained dataset, it is higher as compared to the coarse grained dataset. Additionally, the ability to robustify a network increases with the increase in adversarial noise. 
\end{abstract}

\squeezeup
\squeezeup

\section{Introduction}
Understanding neural networks is crucial in applications like autonomous vehicles, health care, robotics, for validating and debugging, as well as for building the trust of users \cite{kim2018textual, uzunova2019interpretable}. This paper strives to understand and explain the decisions of deep neural networks by studying the behavior of predicted attributes when adversarial examples are introduced.  We argue that even if no adversaries are being inserted in real world applications, adversarial examples can be exploited for understanding neural networks in their failure modes. Most of the state of the art approaches for interpreting neural networks work by focusing on features to produce saliency maps by considering class specific gradient information \cite{selvaraju2017grad, simonyan2013deep, sundararajan2017axiomatic}, or by finding the part of the image which influences classification the most and removing it by adding perturbations \cite{zeiler2014visualizing, fong2017interpretable}. These approaches reveal the part in the image where there is support to the classification and visualize the performance of known good examples. This tells a little about the boundaries of a class where dubious examples reside. 

However, humans motivate their decisions through semantically meaningful observations. For example, this type of bird has a blue head and red belly so, this must be a painted bunting. Hence, we study changes in the predicted attribute values of samples under mild modification of the image through adversarial perturbations. We believe this alternative dimension of study can provide a better understanding of how misclassification in a deep network can best be communicated to humans. Note that, we consider adversarial examples that are generated to fool only the classifier and not the interpretation (attributes) mechanism.    

Interpreting deep neural network decisions for adversarial examples helps in understanding their internal functioning \cite{tao2018attacks,du2018towards}. Therefore, we explore
\begin{figure}[t]
        \includegraphics[width=\linewidth, trim=0 0 0 0, clip]{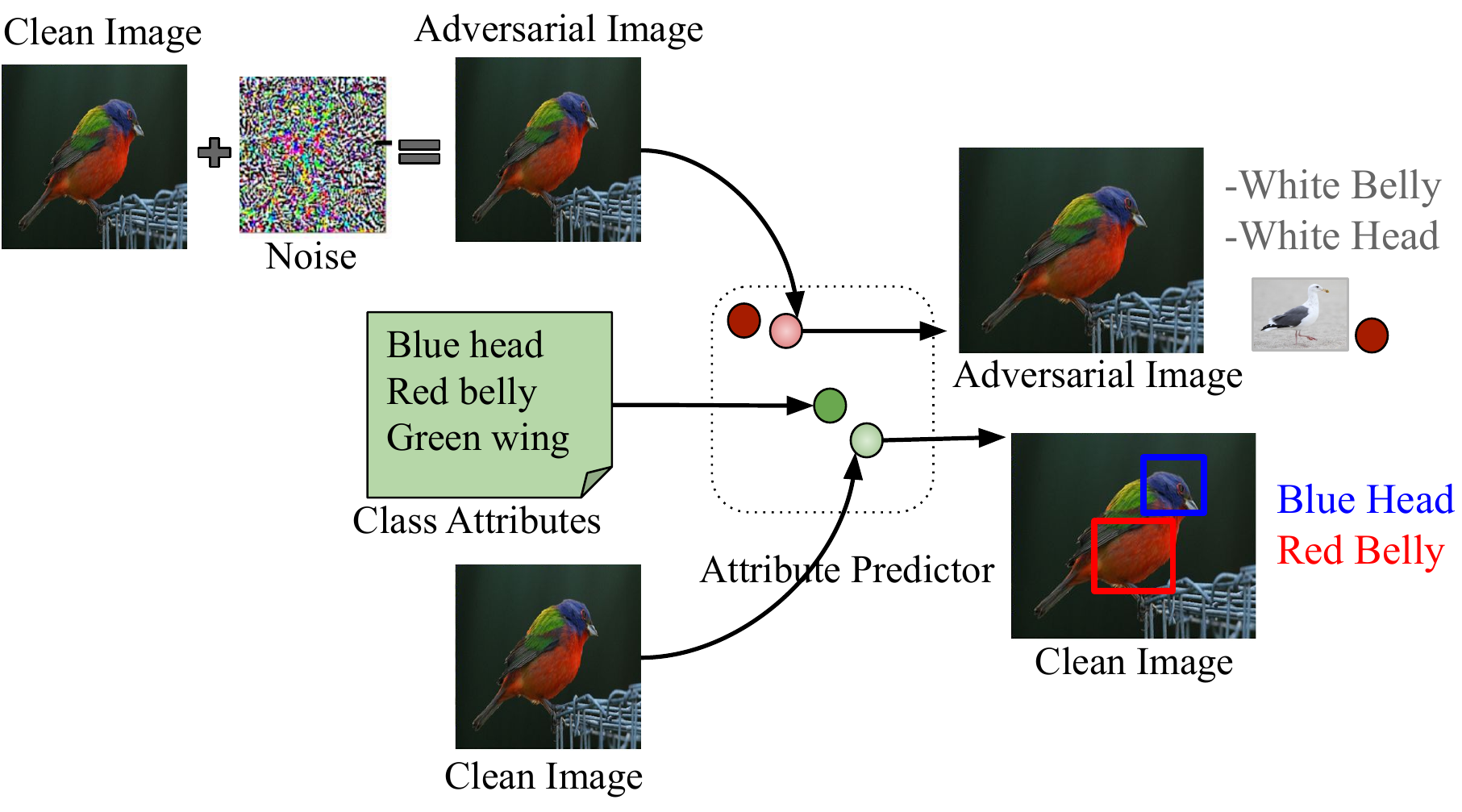}
    \caption{
    Our study with interpretable attribute prediction-grounding framework shows that, for a clean image predicted attributes ``red belly'' and ``blue head'' are coherent with the ground truth class (painted bunting), and for an adversarial image ``white belly'' and ``white head'' are coherent with the wrong class (herring gull) .}
    \label{fig:Motivation}
    \vspace{-3mm}
\end{figure}

\textit{How do the attribute values change under an adversarial attack on the standard classification network?}

However while, describing misclassifications due to adversarial examples with attributes helps in understanding neural networks, assessing whether the attribute values still retain their discriminative power after making the network robust to adversarial noise is equally important. Hence, we also ask

\textit{How do the attribute values change under an adversarial attack on a robust classification network?}

To answer these questions, we design experiments to investigate which attribute values change when an image is misclassified with increasing adversarial perturbations, and further when the classifier is made robust against an adversarial attack. Through these experiments we intend to demonstrate what attributes are important to distinguish between the right and the wrong class. For instance, as shown in Figure~\ref{fig:Motivation}, ``blue head'' and ''red belly'' associated with the class ``painted bunting'' are predicted correctly for the clean image. On the other hand, due to predicting attributes incorrectly as ``white belly'' and ``white head'', the adversarial image gets classified into ``herring gull'' incorrectly. After analysing the changes in attributes with a standard and with a robust network we propose a metric to quantify the robustness of the network against adversarial attacks. Therefore, we ask

\textit{Can we quantify the robustness of an adversarially robust network? }

In order to answer the third question, we design a robustness quantification metric for both standard as well as attribute based classifiers.

To the best of our knowledge we are the first to exploit adversarial examples with attributes to perform a systematic investigation on neural networks, both \textit{quantitatively} and \textit{qualitatively}, for not only \textit{standard}, but also for \textit{adversarially robust} networks. We explain the decisions of deep computer vision systems by identifying what attributes change when an image is perturbed in order for a classification system to produce a specific output.  Our results on three benchmark attribute datasets with varying size and granularity elucidate why adversarial images get misclassified, and why the same images are correctly classified with the adversarially robust framework. Finally we introduce a new metric to quantify the robustness of a network for both general as well as attribute based classifiers.
\squeezeup
\section{Related Work}
In this section, we discuss related work on interpretability and adversarial examples.

\myparagraph{Interpretability.}
Explaining the output of a decision maker is motivated by the need to build user trust before deploying them into the real world environment. Previous work is broadly grouped into two: 1) \textit{rationalization}, that is, justifying the network's behavior and 2) \textit{introspective explanation}, that is, showing the causal relationship between input and the specific output \cite{du2018techniques}. Text-based class discriminative explanations~\cite{hendricks2016generating,park2016attentive}, text-based interpretation with semantic information~\cite{dong2017improving} and counter factual visual explanations \cite{goyal2019counterfactual} fall in the first category. On the other hand activation maximization \cite{simonyan2013deep, zintgraf2017visualizing}, learning the perturbation mask \cite{fong2017interpretable}, learning a model locally around its prediction and finding important features by propagating activation differences \cite{ribeiro2016should,shrikumar2017learning} fall in the second group. The first group has the benefit of being human understandable, but it lacks the causal relationship between input and output. The second group incorporates internal behavior of the network, but lacks human understandability. In this work, we incorporate human understandable justifications through attributes and causal relationship between input and output through adversarial attacks.

\myparagraph{Interpretability of Adversarial Examples.} After analyzing neuronal activations of the networks for adversarial examples in \cite{dong2017towards} it was concluded that the networks learn recurrent discriminative parts of objects instead of semantic meaning. In \cite{jiang2018recent}, the authors proposed a datapath visualization module consisting of the layer level, feature level, and the neuronal level visualizations of the network for clean as well as adversarial images. 
In \cite{zhang2019interpreting}, the authors investigated adversarially trained convolutional neural networks by constructing images with different textural transformations while preserving the shape information to verify the shape bias in adversarially trained networks compared with standard networks. Finally, in \cite{tsipras2018robustness}, the authors showed that the saliency maps from adversarially trained networks align well with human perception.

\begin{figure}[t]
    \centering
        \includegraphics[width=\linewidth, trim=0 0 0 0, clip]{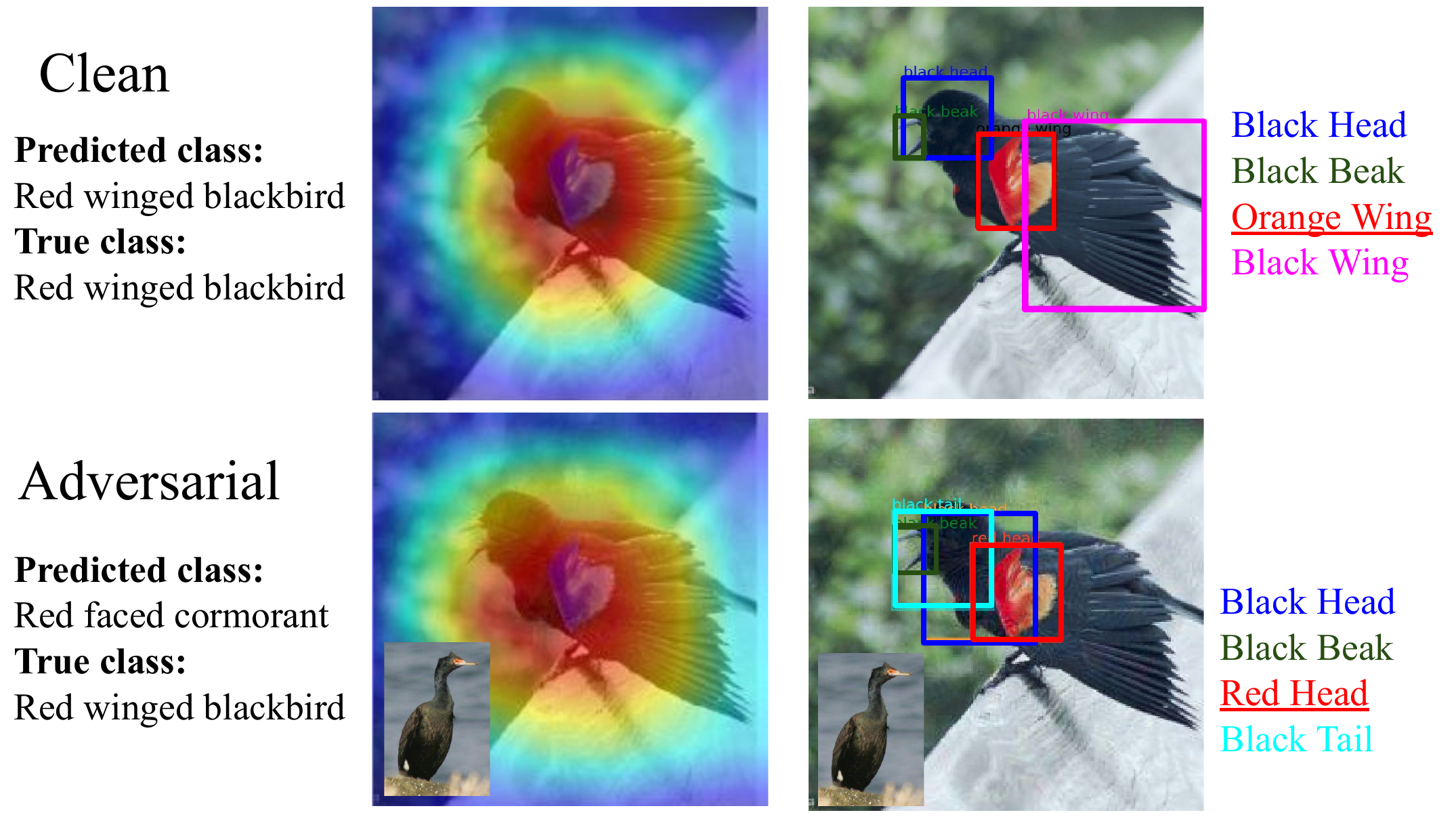}
    \caption{\textbf{Adversarial images are difficult to explain:} when the answer is wrong, often saliency based methods (left) fail to detect what went wrong. Instead, attributes (right) provide intuitive and effective visual and textual explanations.}
   \squeezeup
   \vspace{-1mm}
    \label{fig:saliency}
\end{figure}

These approaches use saliency maps for interpreting the adversarial examples, but 
saliency maps~\cite{selvaraju2017grad} are often weak in justifying classification decisions, especially for fine-grained adversarial images. For instance, in Figure~\ref{fig:saliency} the saliency map of a clean image classified into the ground truth class, ``red winged blackbird'', and the saliency map of a misclassified adversarial image, look quite similar. Instead, we propose to predict and ground attributes for both clean and adversarial images to provide visual as well as attribute-based interpretations. In fact, our predicted attributes for clean and adversarial images look quite different. By grounding the predicted attributes one can infer that the ``orange wing'' is important for ``red winged blackbird'' while the ``red head'' is important for ``red faced cormorant''. Indeed, when the attribute value for orange wing decreases and red head increases the image gets misclassified.

\myparagraph{Adversarial Examples.} Small carefully crafted perturbations, called \textit{adversarial perturbations}, when added to the inputs of deep neural networks, result in \textit{adversarial examples}. These adversarial examples can easily drive the classifiers to the wrong classification~\cite{szegedy2013intriguing}. Such attacks involve iterative fast gradient sign method (IFGSM) \cite{kurakin2016adversarial}, Jacobian-based saliency map attacks \cite{papernot2016limitations}, one pixel attacks \cite{su2019one}, Carlini and Wagner attacks \cite{carlini2017towards} and universal attacks \cite{moosavi2016deepfool}.  We select IFGSM for our experiments, but our method can also be used with other types of adversarial attacks.

Adversarial examples can also be used for understanding neural networks. \cite{anonymous2020evaluations} aims at utilizing adversarial examples for understanding deep neural networks by extracting the features that provide the support for classification into the target class. The most salient features in the images provide the way to interpret the decision of a classifier, but they lack human understandability. Additionally, finding the most salient features is computationally rather expensive. The crucial point, however, is that if humans explain classification by attributes, they are also natural candidates to study misclassification and robustness. Hence, in this work in order to understand neural networks we utilize adversarial examples with attributes which explain the misclassification due to adversarial attacks.

\vspace{-1.5mm}
\section{Method}

In this section, in order to explain what attributes change when an adversarial attack is performed on the classification mechanism of the network, we detail a two-step framework. First, we perturb the images using  adversarial attack methods and robustify the classifiers via adversarial training. Second, we predict the class specific attributes and visually ground them on the image to provide an intuitive justification of why an image is classified as a certain class. Finally, we introduce our metric for quantifying the robustness of an adversarially robust network against adversarial attacks.
  

\subsection{Adversarial Attacks and Robustness}
 Given a clean $n\text{-th}$ input $x_n$ and its respective ground truth class $y_n$ predicted by a model $f(x_n)$, an  adversarial attack model generates an image $\hat{x}_n$ for which the predicted class is $y$, where $y \neq y_n$.  In the following, we detail an adversarial attack method for fooling a general classifier and an adversarial training technique that robustifies it.

\myparagraph{Adversarial Attacks.} The iterative fast gradient sign method (IFGSM)~\cite{kurakin2016adversarial} is leveraged to fool only the classifier network. IFGSM solves the following equation to produce adversarial examples:
\vspace{-1.5mm}
    \begin{align}
        & \hat{x}^0 =x_n \nonumber \\
        & \hat{x}_n^{i+1}=\text{Clip}_{\epsilon}\{\hat{x}_n^{i}+\alpha\text{Sign}(\bigtriangledown{\hat{x}_n^i}\mathcal{L}(\hat{x}_n^i,y_{n}))\}
    \end{align}
where $\bigtriangledown{\hat{x}_n^i}\mathcal{L}$ represents the gradient of the cost function w.r.t. perturbed image $\hat{x}_n^i$ at step $i$. $\alpha$ determines the step size which is taken in the direction of sign of the gradient and finally, the result is clipped by epsilon $\text{Clip}_{\epsilon}$.

\myparagraph{Adversarial Robustness.}  We use \textit{adversarial training} as a defense against adversarial attacks which minimizes the following objective \cite{43405}:
    \begin{align}
        \mathcal{L}_{adv}(x_n,y_n) & = \alpha \mathcal{L}(x_n,y_n)
         + (1-\alpha)\mathcal{L}(\hat{x}_n,y)
    \end{align}
where, $\mathcal{L}(x_n,y_n)$ is the classification loss for clean images, $\mathcal{L}(\hat{x}_n,y)$ is the loss for adversarial images and $\alpha$ regulates the loss to be minimized. The model finds the worst case perturbations and fine tunes the network parameters to reduce the loss on perturbed inputs. Hence, this results in a robust network $f^r(\hat{x})$, using which improves the classification accuracy on the adversarial images.

 \begin{figure}[t]
    \centering
    \includegraphics[width=\linewidth]{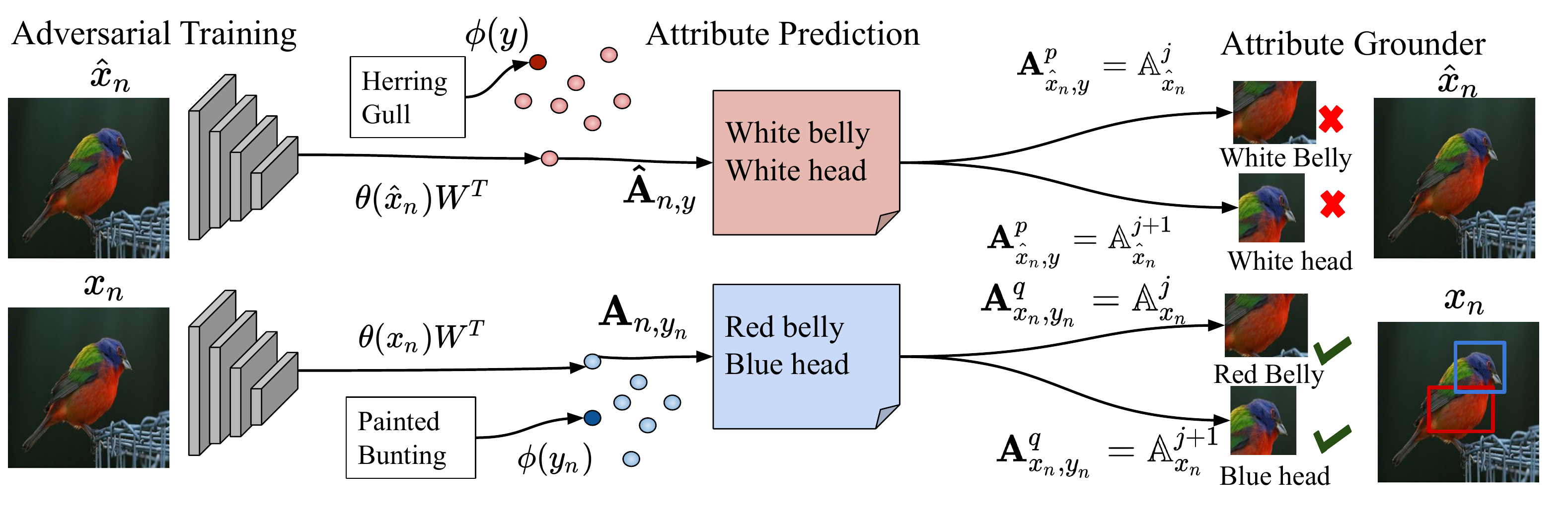}
    \caption{\textbf{Interpretable attribute prediction-grounding model.} After an adversarial attack or adversarial training step, image features of both clean $\theta(x_n)$ and adversarial images $\theta(\hat{x})$ are extracted using Resnet and mapped into attribute space $\phi(y)$ by learning the compatibility function $F(x_n,y_n;W)$ between image features and class attributes. Finally, attributes predicted by attribute based classifier $\bold{A}_{x_n,y_n}^q$ are grounded by matching them with attributes predicted by Faster RCNN $\mathbb{A}_{x_n}^j$ for clean and adversarial images.}
    \squeezeup
    \vspace{-1.5mm}
    \label{fig:ADV_SJE}
\end{figure}  
\vspace{-2mm}
\subsection{Attribute Prediction and Grounding}
Our attribute prediction and grounding model uses attributes to define a joint embedding space that the images are mapped to.

\myparagraph{Attribute prediction.} The model is shown in Figure~\ref{fig:ADV_SJE}. During training our model maps clean training images close to their respective class attributes, e.g. ``painted bunting'' with attributes ``red belly, blue head'', whereas adversarial images get mapped close to a wrong class, e.g. ``herring gull'' with attributes ``white belly, white head''. 

We employ structured joint embeddings (SJE)~\cite{akata2015evaluation} to predict attributes in an image. Given the input image features $\theta(x_n) \in \mathcal{X}$ and output class attributes $\phi(y_n) \in \mathcal{Y}$ from the sample set $\mathcal{S}=\{(\theta(x_n),\phi(y_n),n=1...N \}$  SJE learns a mapping $\mathbb{f}:\mathcal{X} \to \mathcal{Y}$ by minimizing the empirical risk of the form $\frac{1}{N}\sum_{n=1}^N \Delta(y_n,\mathbb{f}(x_n))$ where $\Delta: \mathcal{Y} \times \mathcal{Y} \to \mathbb{R} $ estimates the cost of predicting $\mathbb{f}(x_n)$ when the ground truth label is $y_n$.
    
A compatibility function $F:\mathcal{X}\times\mathcal{Y}\to \mathbb{R}$ is defined between input $\mathcal{X}$and output $\mathcal{Y}$ space:
\begin{equation}
        F(x_n,y_n;W)=\theta(x_n)^TW\phi(y_n)
\end{equation}
Pairwise ranking loss  $\mathbb{L}(x_n,y_n,y)$ is used to learn the parameters $(W)$:
    \begin{equation}
        \Delta(y_n,y)+\theta(x_n)^TW\phi(y_n)-\theta(x_n)^TW\phi(y)
    \end{equation}
Attributes are predicted for both clean and adversarial images by:
\vspace{-3mm}
\begin{equation}
\vspace{-3mm}
    \bold{A}_{n,y_n}=\theta(x_n)W \, , \bold{\hat{A}}_{n,y}=\theta(\hat{x}_n)W 
\end{equation}
The image is assigned to the label of the nearest output class attributes $\phi(y_n)$. 

\myparagraph{Attribute grounding.} In our final step, we ground the predicted attributes on to the input images using a pre-trained Faster RCNN network and visualize them as in~\cite{anne2018grounding}. The pre-trained Faster RCNN model $\mathcal{F}(x_n)$ predicts the bounding boxes denoted by $b^j$. For each object bounding box it predicts the class $\mathbb{Y}^j$ as well as the attribute $\mathbb{A}^j$ ~\cite{anderson2018bottom}.
\begin{equation}
\vspace{-1.5mm}
    b^j,\mathbb{A}^j,\mathbb{Y}^j=\mathcal{F}(x_n)
\end{equation}

where, $j$ is the bounding box index. The most discriminative attributes predicted by SJE are selected based on the criteria that they change the most when the image is perturbed with noise. For clean images we use: 
\begin{equation}
        q=\underset{i}{\mathrm{argmax}}(\bold{A}_{n,y_n}^i-\phi(y^i))
        \label{eq:att_sel1}
        \vspace{-1mm}
\end{equation}
and for adversarial images we use: 
\begin{equation}
        p=\underset{i}{\mathrm{argmax}}(\bold{\hat{A}}_{n,y}^i-\phi(y_n^i)). 
        \label{eq:att_sel2}
         \vspace{-1mm}
\end{equation}
where $i$ is the attribute index, $q$ and $p$ are the indexes of the most discriminative attributes predicted by SJE and $\phi(y^i)$, $\phi(y_n^i)$ are wrong class and ground truth class attributes respectively. Then we search for selected attributes $\bold{A}_{x_n,y_n}^q, \bold{A}_{\hat{x}_n,y}^p$ in attributes predicted by Faster RCNN for each bounding box $\mathbb{A}_{x_n}^j, \mathbb{A}_{\hat{x}_n}^j$, and when the attributes predicted by SJE and Faster RCNN are found, that is $\bold{A}_{x_n,y_n}^q = \mathbb{A}_{x_n}^j$, $\bold{A}_{\hat{x}_n,y}^p = \mathbb{A}_{\hat{x}_n}^j$ we ground them on their respective clean and adversarial images. Note that the adversarial images being used here are generated to fool only the general classifier \textit{and not the attribute predictor nor the Faster RCNN}.

\subsection{Robustness Quantification}
To describe the ability of a network for robustification, independent of its performance on a standard classifier we introduce a metric called \textit{robust ratio}. We calculate the loss of accuracy $L_R$ on a robust classifier, by comparing a standard classifier $f(x_n)$ on clean images with the robust classifier $f^r(\hat{x}_n)$ on the adversarially perturbed images as given below: 
\begin{equation}
    L_R=f(x_n)-f^r(\hat{x}_n)
\end{equation}
And then we calculate the loss of accuracy $L_S$ on a standard classifier, by comparing its accuracy on the clean and adversarially perturbed images:
\vspace{-2mm}
\begin{equation}
    L_S=f(x_n)-f(\hat{x}_n)
\end{equation}
The ability to robustify is then defined as: 
\vspace{-2mm}
\begin{equation}
    R=\frac{L_R}{L_S}
\end{equation}
$R$ is the robust ratio. It indicates the fraction of the classification accuracy of the standard classifier recovered by the robust classifier when adding noise.

\vspace{-1.5mm}
\section{Experiments}

     \begin{figure*}[t]
        \centering
            \includegraphics[width=0.324\linewidth, trim=15 0 45 20, clip]{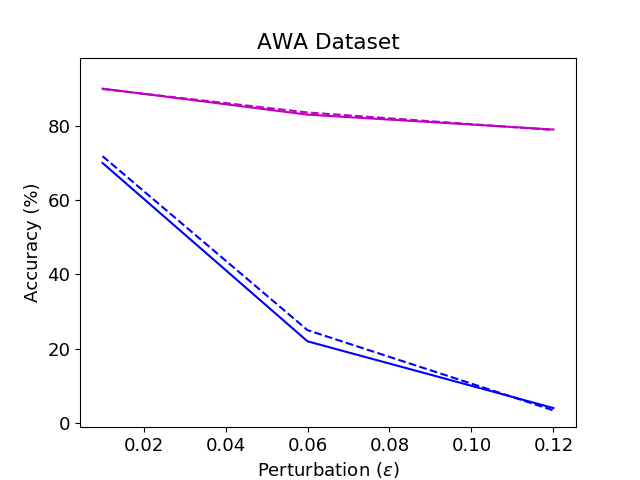}
            \includegraphics[width=0.324\linewidth, trim=15 0 45 20, clip]{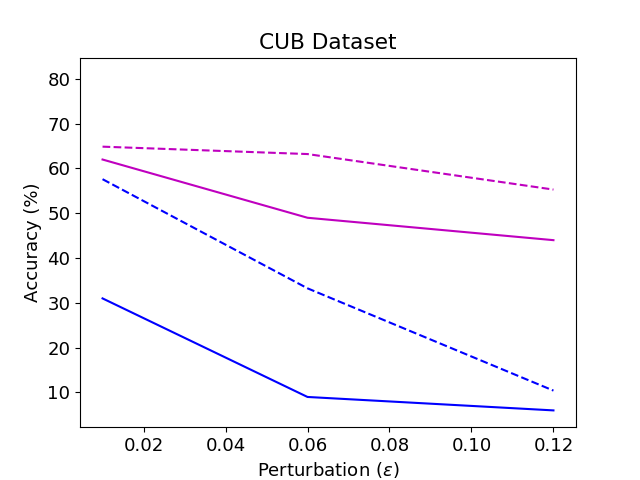} 
            \includegraphics[width=0.324\linewidth, trim =10 115 0 110, clip]{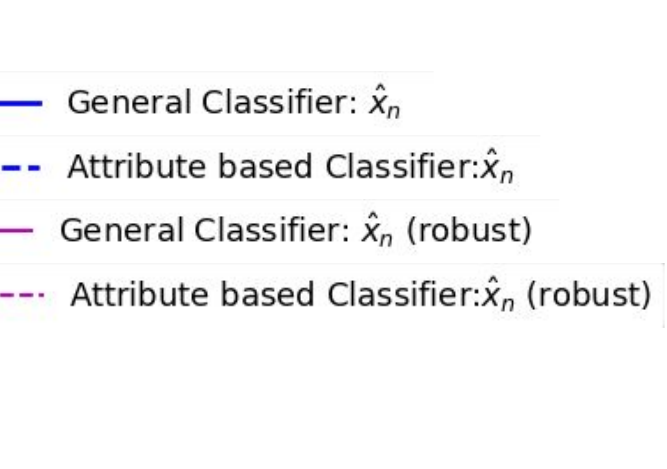}
        \caption{\textbf{Comparing the accuracy of the general and the attribute based classifiers for adversarial examples to investigate change in attributes.} We evaluate both classifiers by extracting features from a standard network and the adversarially robust network.}
        \squeezeup
        \label{fig:acc_plots}
    \end{figure*}
In this section, we perform experiments on three different datasets and analyse the change in attributes for clean as well as adversarial images. We additionally analyse results for our proposed robustness quantification metric on both general and attribute based classifiers.

\myparagraph{Datasets.} We experiment on three datasets, Animals with Attributes 2 (AwA) \cite{lampert2009learning}, Large attribute (LAD) \cite{zhao2018large} and Caltech UCSD Birds (CUB)  \cite{wah2011caltech}. AwA contains 37322 images (22206 train / 5599 val / 9517 test) with 50 classes and 85 attributes per class. LAD has 78017 images (40957 train / 13653 val / 23407 test) with 230 classes and 359 attributes per class. CUB consists of 11,788 images (5395 train / 599 val / 5794 test) belonging to 200 fine-grained categories of birds with 312 attributes per class. All the three datasets contain real valued class attributes representing the presence of a certain attribute in a class. 

Visual Genome Dataset \cite{krishna2017visual} is used to train the Faster-RCNN model which extracts the bounding boxes using 1600 object and 400 attribute annotations. Each bounding box is associated with an attribute followed by the object, e.g. a brown bird.

\myparagraph{Image Features and Adversarial Examples.} We extract image features and generate adversarial images using the fine-tuned Resnet-152. Adversarial attacks are performed using IFGSM method with epsilon $\epsilon$ values $0.01$, $0.06$ and $0.12$. The $\l_\infty $ norm is used as a similarity measure between clean input and the generated adversarial example. 
 
 \myparagraph{Adversarial Training.}
As for adversarial training, we repeatedly computed the adversarial examples while training the fine-tuned Resnet-152 to minimize the loss on these examples. We generated adversarial examples using the projected gradient descent method. This is a multi-step variant of FGSM with epsilon $\epsilon$ values $0.01$, $0.06$ and $0.12$ respectively for adversarial training as in~\cite{madry2017towards}.  

Note that we are not attacking the attribute based network directly but we are attacking the general classifier and extracting features from it for training the attribute based classifier. Similarly, the adversarial training is also performed on the general classifier and the features extracted from this model are used for training the attribute based classifier.

\myparagraph{Attribute Prediction and Grounding.}

At test time the image features are projected onto the attribute space. The image is assigned with the label of the nearest ground truth attribute vector. The predicted attributes are grounded by using Faster-RCNN pre-trained on Visual Genome Dataset since we do not have ground truth part bounding boxes for any of attribute datasets. 
\squeezeup
\section{Results}
We investigate the change in attributes quantitatively (i) by performing classification based on attributes and (ii) by computing distances between attributes in embedding space. We additionally investigate changes qualitatively by grounding the attributes on images for both standard and adversarially robust networks. 

At first, we compare the general classifier $f(x_n)$ and the attribute based classifier $\mathbb{f}(x_n)$ in terms of the classification accuracy on clean images. Since the attribute based model is a more explainable classifier, it predicts attributes, compared to general classifier, which predicts the class label directly. Therefore, we first verify whether the attribute based classifier performs equally well as the general classifier. We find that, the attribute based and general classifier accuracies are comparable for AWA (general: 93.53, attribute based: 93.83). The attribute based classifier accuracy is slightly higher for LAD (general: 80.00, attribute based: 82.77), and slightly lower for CUB (general: 81.00, attribute based: 76.90) dataset.
 \begin{figure}[t]
        \centering
        \includegraphics[width=\linewidth, trim=0 0 0 30, clip]{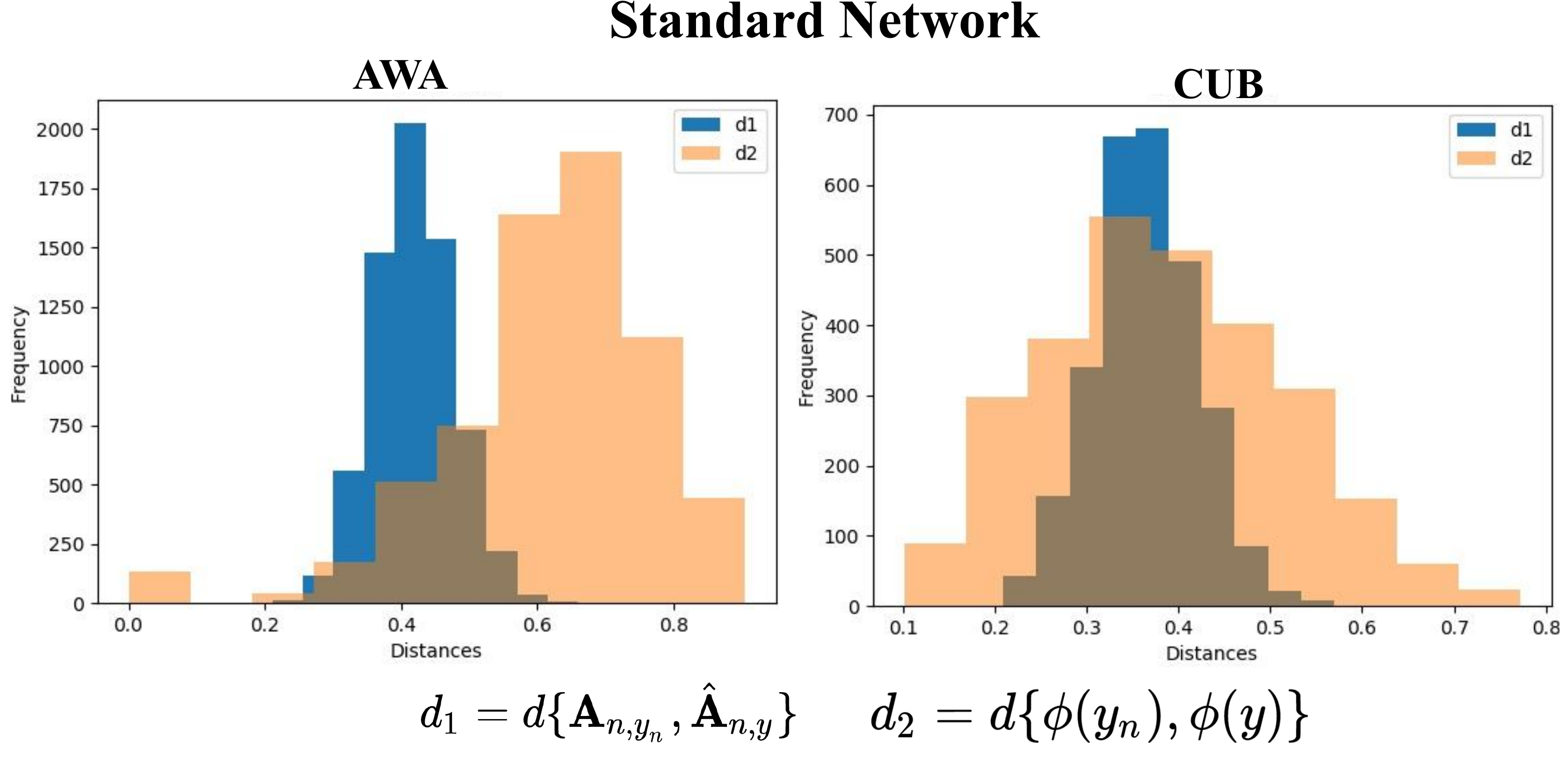}
        \vspace{-3mm}
        \caption{ \textbf{Attribute distance plots for standard learning frameworks.} Standard learning framework plots are shown for the clean and the adversarial image attributes.}
            \label{fig:standardattr}
            \vspace{-3mm}
\end{figure}

\begin{figure*}[t]
    \centering
    \includegraphics[width=\linewidth, trim=0 5 0 30, clip]{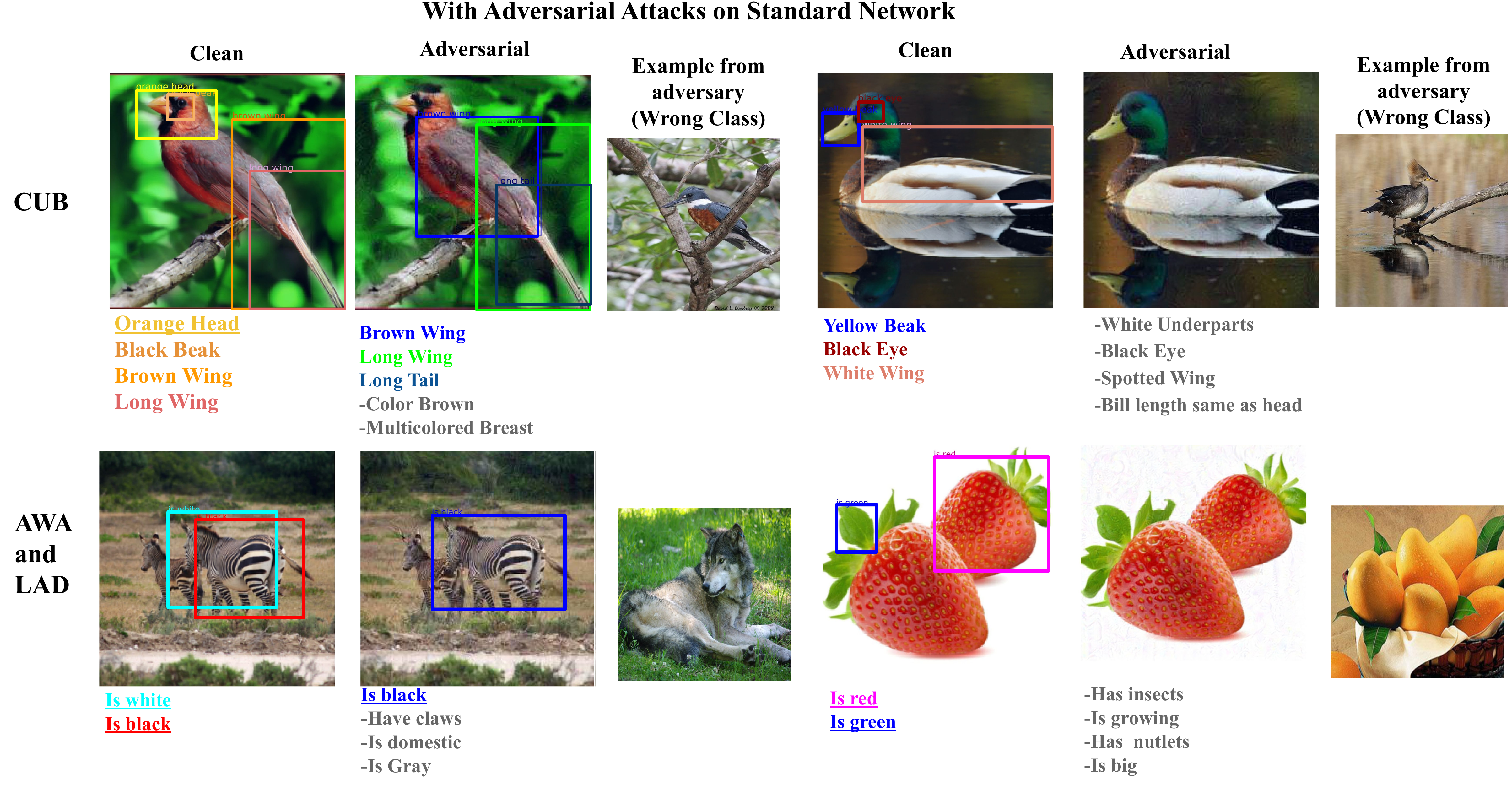}
    \vspace{-6mm}
    \caption{\textbf{Qualitative analysis for  adversarial attacks on standard network.} The attributes ranked by importance for the classification decision are shown below the images. The grounded attributes are color coded for visibility (the ones in gray could not be grounded). The attributes for clean images are related to the ground truth classes whereas the ones predicted for adversarial images are related to the wrong classes. }
    \label{fig:Qualitative-1}
    \vspace{-3mm}
\end{figure*} 
To qualitatively analyse the predicted attributes, we ground them on clean and adversarial images. We select our images among the ones that are correctly classified when clean and incorrectly classified when adversarially perturbed. Further we select the most discriminative attributes based on equation \ref{eq:att_sel1} and \ref{eq:att_sel2}. We evaluate $50$ attributes that change their value the most for the CUB, $50$ attributes for the AWA, and  $100$ attributes for the LAD dataset.
\subsection{Adversarial Attacks on Standard Network}
\subsubsection{Quantitative Analysis}
\squeezeup
 \myparagraph{By Performing Classification based on Attributes.} With adversarial attacks, the accuracy of both the general and attribute based classifiers drops with the increase in perturbations see Figure~\ref{fig:acc_plots} (blue curves). The drop in accuracy of the general classifier for the fine grained CUB dataset is higher as compared to the coarse AWA dataset which confirms our hypothesis. For example, at $\epsilon=0.01$ for the CUB dataset the general classifier's accuracy drops from $81\%$ to $31\%$ ($\approx 50\%$ drop), while for the AWA dataset it drops from $93.53\%$ to $70.54\%$ ($\approx 20\%$ drop). However, the drop in accuracy with the attribute based classifier is almost equal for both, $\approx 20\%$ . We propose one of the reasons behind the smaller drop of accuracy for the CUB dataset with the attribute based classifier compared to the general classifier is that for fine grained datasets there are many common attributes among classes. Therefore, in order to misclassify an image a significant number of attributes need to be changed. For a coarse grained dataset, changing a few attributes is sufficient for misclassification. Another reason is that there are $9\%$ more attributes per class in the CUB dataset as compared to the AWA dataset.
\vspace{-0.5mm}

For the coarse dataset the attribute based classifier shows comparable performance with the general classifier. While for the fine grained dataset the attribute based classifier shows better performance than the general classifier so a large change in attributes is required to cause misclassification with attributes. Overall, the drop in the accuracy with the adversarial attacks demonstrates that, with adversarial perturbations, the attribute values change towards those that belong to the new class and cause the misclassification.

\myparagraph{By Computing Distances in Embedding Space.} In order to perform analysis on attributes in embedding space, we consider the images which are correctly classified without perturbations and misclassified with perturbations. Further, we select the top $20\%$ of the most discriminative attributes using equation \ref{eq:att_sel1} and \ref{eq:att_sel2}. Our aim is to analyse the change in attributes in embedding space.

\begin{figure}[t]
        \centering
        \includegraphics[width=\linewidth, trim=0 0 0 27, clip]{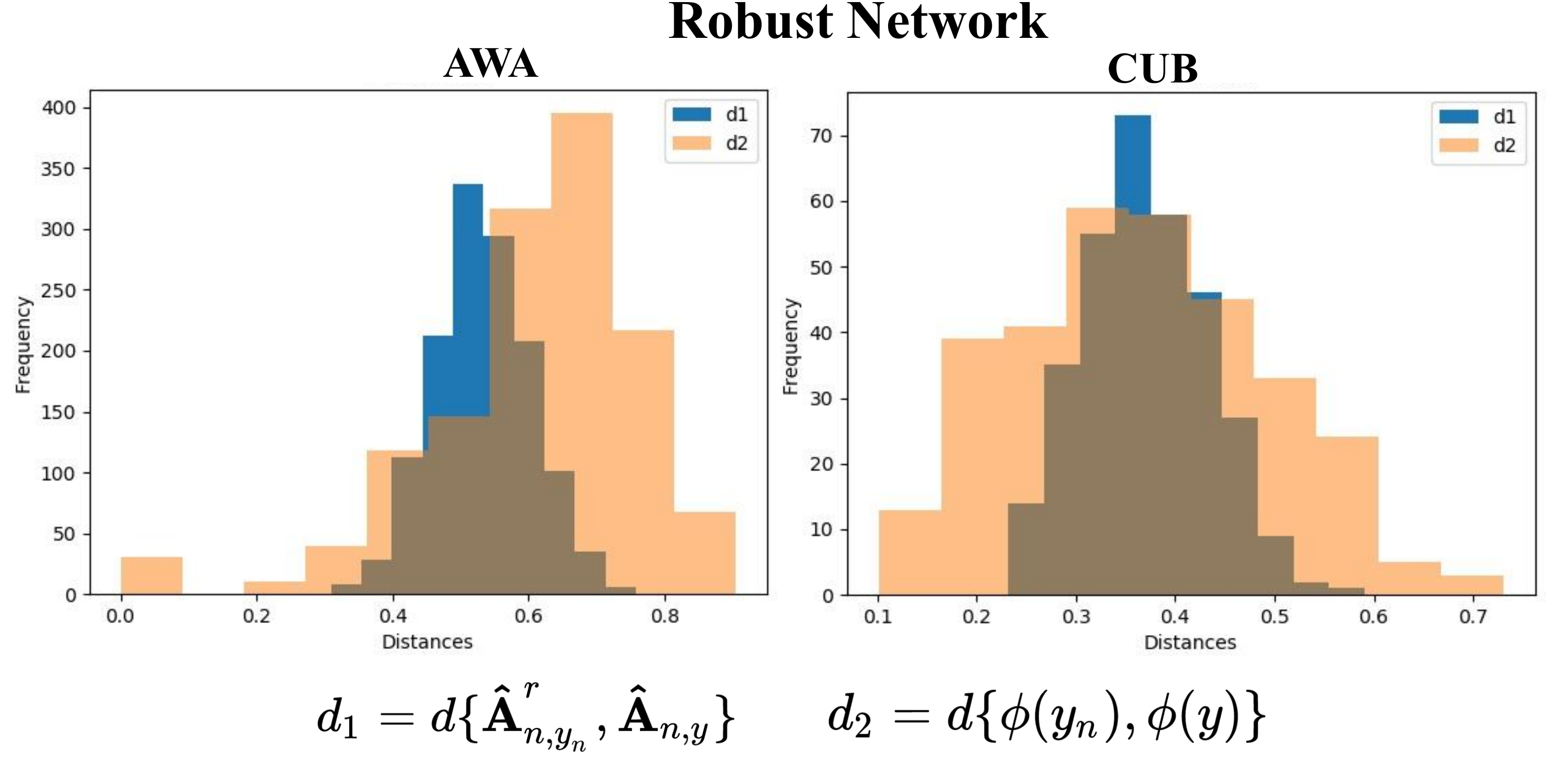}
        \vspace{-3mm}
        \caption{ \textbf{Attribute distance plots for robust learning frameworks.} Robust learning framework plots are shown only  for  the adversarial image attributes but for adversarial images misclassified with the standard features and correctly classified with the robust features.}
            \label{fig:robustattr}
            \vspace{-3mm}
            \squeezeup
\end{figure}
We contrast the Euclidean distance between predicted attributes of clean and adversarial samples:
\begin{equation}
    d_1 = d\{\bold{A}_{n,y_n},\bold{\hat{A}}_{n,y}\} =\parallel \bold{A}_{n,y_n}-\bold{\hat{A}}_{n,y} \parallel_2
    \label{eq:d1_1}
\end{equation}
with the Euclidean distance between the ground truth attribute vector of the correct and wrong classes:
\begin{equation}
   d_2 = d\{\phi(y_n),\phi(y)\}=\parallel\phi(y_n)-\phi(y)) \parallel_2
   \label{eq:d2_1}
\end{equation}
and show the results in Figure~\ref{fig:standardattr}. Where, $\bold{A}_{n,y_n}$ denotes the predicted attributes for the clean images classified correctly, and $\bold{\hat{A}}_{n,y}$ denotes the predicted attributes for the adversarial images misclassified with a standard network. The correct ground truth class attribute is referred to as $\phi(y_n)$ and wrong class attributes are $\phi(y)$.

\begin{figure*}[t]
    \centering
    \includegraphics[width=\linewidth, trim=0 0 0 30, clip]{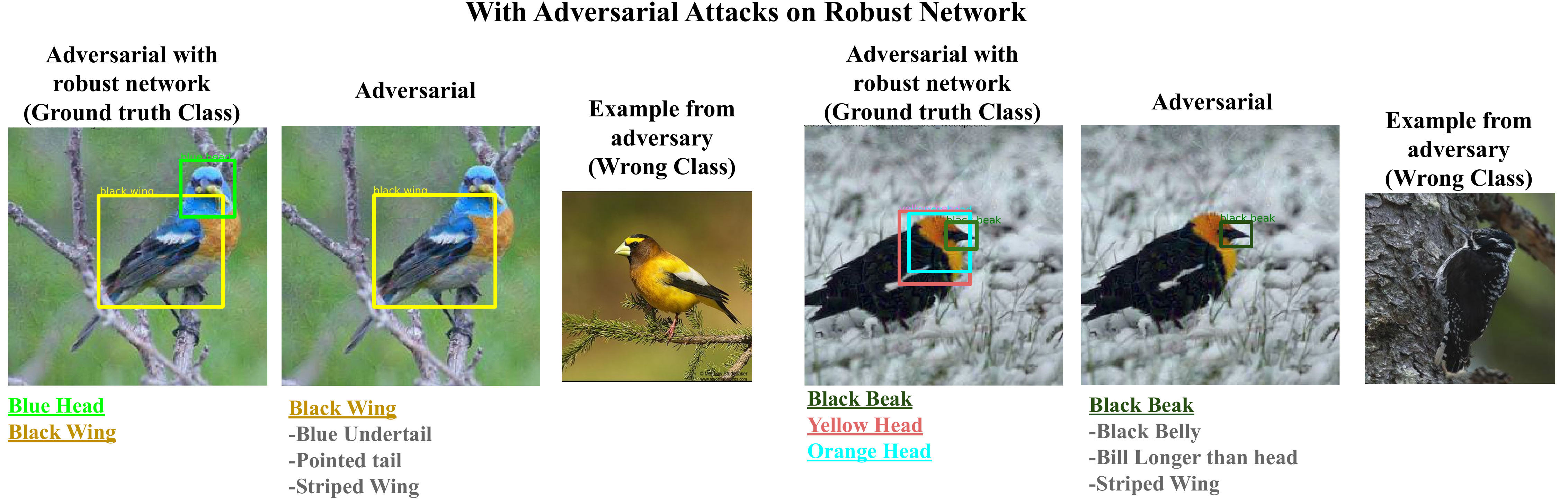}
    \vspace{-6mm}
    \caption{\textbf{Qualitative analysis for adversarial attacks on robust network.} The attributes are ranked by importance for the classification decision, the grounded attributes are color coded for visibility (the ones in gray could not be grounded). The attributes for adversarial images with robust network are related to ground truth classes whereas the ones predicted for adversarial images change towards wrong classes }
    \label{fig:Qualitative-2}
    \vspace{-3mm}
    
\end{figure*} 

We observe that for the AWA dataset the distances between the predicted attributes for adversarial and clean images $d_1$ are smaller than the distances between the ground truth attributes of the respective classes $d_2$. The closeness in predicted attributes for clean and adversarial images as compared to their ground truths shows that attributes change towards the wrong class but not completely. This is due to the fact that for coarse classes, only a small change in attribute values is sufficient to change the class.

The fine-grained CUB dataset behaves differently. The overlap between $d_1$ and $d_2$ distributions demonstrates that attributes of images belonging to fine-grained classes change significantly as compared to images from coarse categories. Although the fine grained classes are closer to each other, due to the existence of many common attributes among fine grained classes, attributes need to change significantly to cause misclassification. Hence, for the coarse dataset, the attributes change minimally, while for the fine grained dataset they change significantly.
\squeezeup
\squeezeup
\subsubsection{Qualitative Analysis}
We observe in Figure~\ref{fig:Qualitative-1} that the most discriminative attributes for the clean images are coherent with the ground truth class and are localized accurately; however, for adversarial images they are coherent with the wrong class. Those attributes which are common among both clean and adversarial classes are localized correctly on the adversarial images; however, the attributes which are not related to the ground truth class, the ones that are related to the wrong class can not get grounded as there is no visual evidence that supports the presence of these attributes. For example ``brown wing, long wing, long tail'' attributes are common in both classes; hence, they are present both in the clean image and the adversarial image. On the other hand, ``has a brown color'' and ``a multicolored breast'' are related to the wrong class and are not present in the adversarial image. Hence, they can not be grounded. Similarly, in the second example none of the attributes are grounded. This is because attributes changed completely towards the wrong class and the evidence for those attributes is not present in the image. This indicates that attributes for the clean images correspond to the ground truth class and for adversarial images correspond to the wrong class. Additionally, only those attributes common among both the wrong and the ground truth classes get grounded on adversarial images.

Similarly, our results on the LAD and AWA datasets in the second row of Figure~\ref{fig:Qualitative-1} show that the grounded attributes on clean images confirm the classification into the ground truth class while the attributes grounded on adversarial images are common among clean and adversarial images. For instance, in the first example of AWA, the ``is black'' attribute is common in both classes so it is grounded on both images, but ``has claws'' is an important attribute for the adversarial class. As it is not present in the ground truth class, it is not grounded. 

\begin{figure*}[t]
    \centering
    \includegraphics[width=0.35\linewidth, trim=0 0 10 20, clip]{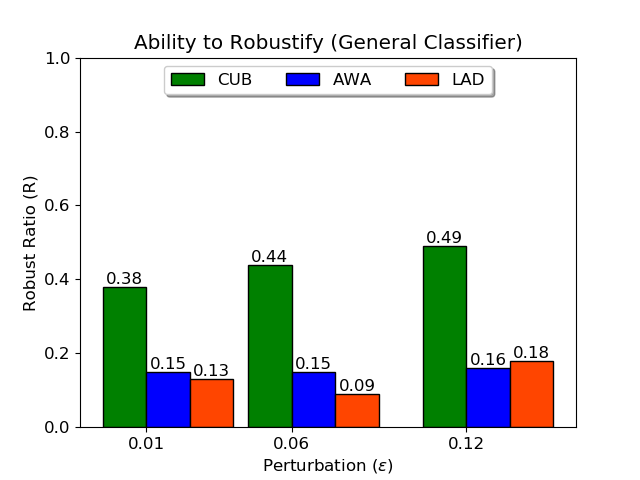}
    \includegraphics[width=0.35\linewidth, trim=0 0 10 20, clip]{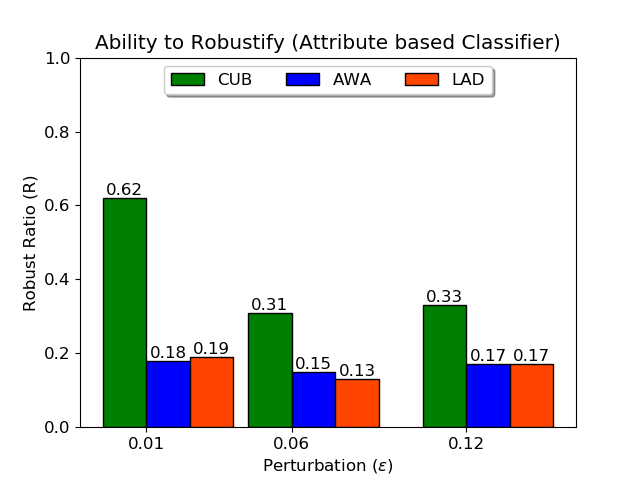}
    \vspace{-1mm}
    \caption{\textbf{Ability to robustify a network.} Ability to robustify a network with increasing adversarial perturbations is shown for three different datasets for both general and attribute based classifiers.}
    \label{fig:Robustifiability}
    \vspace{-4mm}
\end{figure*}

Compared to misclassifications caused by adversarial perturbations on CUB, images do not necessarily get misclassified into the most similar class for the AWA and LAD datasets as they are coarse grained datasets. Therefore, there is less overlap of attributes between ground truth and adversarial classes, which is in accordance with our quantitative results. Furthermore, the attributes for both datasets are not highly structured, as different objects can be distinguished from each other with only a small number of attributes. 

\subsection{Adversarial Attacks on Robust Network}
\subsubsection{Quantitative Analysis}
\squeezeup
\myparagraph{By Performing Classification based on Attributes.} Our evaluation on the standard and adversarially robust networks shows that the classification accuracy improves for the adversarial images when adversarial training is used to robustify the network: \ref{fig:acc_plots} (purple curves). For example, in Figure ~\ref{fig:acc_plots} for AWA the accuracy of the general classifier improved from $ 70.54\%$ to $92.15\%$ ($\approx 21\%$ improvement) for adversarial attack with $\epsilon=0.01$. As expected for the fine grained CUB dataset the improvement is $\approx 31\%$ higher than the AWA dataset. However, for the attribute based classifier, the improvement in accuracy for AWA ($\approx 18.06\%$) is almost double that of the CUB dataset ($\approx 7\%$). We propose this is because the AWA dataset is coarse, so in order to classify an adversarial image correctly to its ground truth class, a small change in attributes is sufficient. Conversely the fine grained CUB dataset requires a large change in attribute values to correctly classify an adversarial image into its ground truth class. Additionally, CUB contains $9\%$ more per class attributes. For a coarse AWA dataset the attributes change back to the correct class and represent the correct class accurately. While for the fine grained CUB dataset, a large change in attribute values is required to correctly classify images.

This shows that with a robust network, the change in attribute values for adversarial images indicate to the ground truth class, resulting in better performance. Overall, we observe by analysing attribute based classifier accuracy that with the adversarial attacks the change in attribute values indicates in which wrong class it is assigned and with the robust network the change in attribute values indicates towards the ground truth class.

\myparagraph{By Computing Distances in Embedding Space}

We compare the distances between the predicted attributes of only adversarial images that are classified correctly with the help of an adversarially robust network $\bold{\hat{A}}^{{r}}_{n,y_n}$ and classified incorrectly with a standard network $\bold{\hat{A}}_{n,y}$:
\squeezeup
\begin{equation}\label{eq:d1_3}
    d_1 = d\{\bold{\hat{A}}^{{r}}_{n,y_n},\bold{\hat{A}}_{n,y}\}=\parallel \bold{\hat{A}}^{{r}}_{n,y_n}-\bold{\hat{A}}_{n,y} \parallel_2
    \squeezeup
\end{equation}
with the distances between the ground truth target class attributes $\phi(y_n)$ and ground truth wrong class attributes $\phi(y)$:
\squeezeup
\begin{equation}\label{eq:d2_3}
    d_2 = d\{\phi(y_n),\phi(y)\}=\parallel\phi(y_n)-\phi(y)) \parallel_2
\end{equation}
The results are shown in Figure~\ref{fig:robustattr}. By comparing Figure~\ref{fig:robustattr} with Figure~\ref{fig:standardattr} we observe a similar behavior. The plots in Figure~\ref{fig:standardattr} are plotted between clean and adversarial image attributes. While plots in Figure~\ref{fig:robustattr} are plotted between only adversarial images but classified correctly with an adversarially robust network and misclassified with a standard network. This shows that the adversarial images classified correctly with a robust network behave like clean images, i.e.  a robust network predicts attributes for the adversarial images which are closer to their ground truth class. 
\squeezeup
\subsubsection{Qualitative Analysis}
 Finally, our analysis with correctly classified images by the adversarially robust network shows that adversarial images with the robust network behave like clean images also visually. In the Figure~\ref{fig:Qualitative-2}, we observe that the attributes of an adversarial image with a standard network are closer to the adversarial class attributes. However, the grounded attributes of adversarial image with a robust network are closer to its ground truth class. For instance, the first example contains a ``blue head'' and a ``black wing'' whereas one of the most discriminating properties of the ground truth class ``blue head'' is not relevant to the adversarial class. Hence this attribute is not predicted as the most relevant by our model, and thus our attribute grounder did not ground it. This shows that the attributes for adversarial images classified correctly with the robust network are in accordance with the ground truth class and hence get grounded on the adversarial images.
 \subsection{Analysis for Robustness Quantification}
 The results for our proposed robustness quantification metric are shown in Figure~\ref{fig:Robustifiability}. We observe that the ability to robustify a network against adversarial attacks varies for different datasets. The network with fine grained CUB dataset is easy to robustify as compared to coarse AWA and LAD datasets. For the general classifier as expected the ability to robustify the network increases with the increase in noise. For the attribute based classifier the ability to robustify the network is high with the small noise but it drops as the noise increases (at $\epsilon=0.06$) and then again increases at high noise value (at $\epsilon=0.12$).

 \squeezeup
\section{Conclusion}

In this work we conducted a systematic study on understanding the neural networks by exploiting adversarial examples with attributes. We showed that if a noisy sample gets misclassified then its most discriminative attribute values indicate to which wrong class it is assigned. On the other hand, if a noisy sample is correctly classified with the robust network then the most discriminative attribute values indicate towards the ground truth class. Finally, we proposed a metric for quantifying the robustness of a network and showed that the ability to robustify a network varies for different datasets. Overall the ability to robustify a network increases with the increase in adversarial perturbations.
{\small
\bibliographystyle{ieee}
\bibliography{egbib}
}

\end{document}